# An Efficient 3D Latent Diffusion Model for T1-contrast Enhanced MRI Generation

Zach Eidex[1], Mojtaba Safari[1], Jie Ding[1,3], Richard L.J. Qiu[1,3], Justin Roper[1,3], David S. Yu[1,3], Hui-Kuo Shu[1,3], Zhen Tian[4], Hui Mao[2,3] and Xiaofeng Yang[1,3*]

[1]Department of Radiation Oncology, Emory University, Atlanta, GA
[2]Department of Radiology and Imaging Sciences, Emory University, Atlanta, GA
[3]Winship Cancer Institute, Emory University, Atlanta, GA
[4]Department of Radiation & Cellular Oncology, University of Chicago, Chicago, IL

**Running title:** T1 Contrast Synthesis

**Manuscript Type:** Original Research

**Contact information:**
Email – xiaofeng.yang@emory.edu
Address - 1365-C Clifton Road NE Atlanta, Georgia 30322

# ABSTRACT

**Objective:** Gadolinium-based contrast agents (GBCAs) are commonly employed with T1-weighted (T1w) MRI to enhance lesion visualization but are restricted in patients at risk of nephrogenic systemic fibrosis and variations in GBCA administration can introduce imaging inconsistencies. This study develops an efficient 3D deep-learning framework to generate T1-contrast enhanced images (T1C) from pre-contrast multiparametric MRI.

**Approach:** We propose the 3D latent rectified flow (T1C-RFlow) model for generating high-quality T1C images. First, T1w and T2-FLAIR images are input into a pretrained autoencoder to acquire an efficient latent space representation. A rectified flow diffusion model is then trained in this latent space representation. The T1C-RFlow model was trained on a curated dataset comprised of the Brain Tumor Segmentation (BraTS) 2024 glioma (GLI; 1480 patients), meningioma (MEN; 1141 patients), and metastases (MET; 1475 patients) datasets. Selected patients were split into training (N=2860), validation (N=612), and test (N=614) sets. Model performance was evaluated with the normalized mean squared error (NMSE) and structural similarity index measure (SSIM).

**Results:** Both qualitative and quantitative results demonstrate that the T1C-RFlow model outperforms benchmark 3D models (pix2pix, denoising diffusion probability models (DDPM), Diffusion Transformers (DiT-3D)) trained in the same latent space. T1C-RFlow achieved the following metrics - GLI: NMSE 0.044 ± 0.047, SSIM 0.935 ± 0.025; MEN: NMSE 0.046 ± 0.029, SSIM 0.937 ± 0.021; MET: NMSE 0.098 ± 0.088, SSIM 0.905 ± 0.082. Further studies showed T1C-RFlow to have the best tumor reconstruction performance and significantly faster denoising times (6.9 s/volume, 200 steps) than conventional DDPM models in both latent space (37.7s, 1000 steps) and patch-based in image space (4.3 hr/volume).

**Significance:** Our proposed method generates synthetic T1C images that closely resemble radiological features of ground truth T1C in much less time than previous diffusion models. Further development may permit a practical method for contrast-agent-free MRI for brain tumors.

# 1. INTRODUCTION

T1-contrast-enhanced MRI (T1C) is frequently used in oncologic, neural, and vascular imaging workflows where accurate lesion delineation is critical for diagnosis and treatment planning. (Eidex et al., 2023, Ibrahim et al., 2018) By injecting a gadolinium-based contrast agent (GBCA, abnormalities such as tumors, inflammation, etc., are significantly enhanced in T1-weighted (T1w) imaging (Roozpeykar et al., 2022). In oncology, the enhanced tumor is defined as the region with significant contrast uptake on T1C images and is useful in classifying the lesion. (Zhou and Lu, 2013) However, GBCAs introduce safety concerns (e.g., nephrogenic systemic fibrosis due to gadolinium retention), extend scanning time, and there can be variations in GBCA administrations due to operational complexity and limited availability and affordability in certain resource limiting regions. (Iyad et al., 2023) Therefore, there is a strong clinical interest in developing contrast agent-free alternatives that can capture diagnostic features and information in T1C images.

Although deep-learning methods have made significant progress in generating synthetic T1C (sT1C), key limitations remain. The majority of studies still operate on 2D axial slices which are memory-efficient but cannot provide sufficient through-plane consistency and long-range 3D context. (Eidex et al., 2025a, Li et al., 2025a) Fully 3D networks improve anatomical coherence but rely on patch-based training and sliding-window inference to fit GPU memory, so long-range relationships are not well captured.(Pan et al., 2024, Shaoyan et al., 2025) Furthermore, constructing the sT1C volume from patches not only leads to long prediction times but also can introduce overlap-stitching and boundary artefacts. (Bieder et al., 2024)

Denoising diffusion probability models (DDPMs) (Ho et al., 2020, Chang et al., 2024) are the current state-of-the-art method over generational adversarial networks (GANs) like pix2pix (Isola et al., 2018) since they deliver strong fidelity without the training stability challenges of GANs. (Becker et al., 2022) However, DDPMs require one thousand steps to reconstruct the image from noise and input priors (T1w and T2-FLAIR MRI) , so high-resolution inference times can take several hours, limiting applicability to clinical practice. To overcome this issue, latent diffusion models (LDMs) (Rombach et al., 2022) first compress the input volumes into a compressed latent space representation, so the computational requirements and inference times are significantly lessened. Furthermore, advanced noise schedulers like the denoising diffusion implicit model (DDIM) (Song et al., 2022), pseudo numerical methods for diffusion model (PNDM) (Liu et al., 2022a), and rectified flow (RFlow) (Esser et al., 2024) schedulers reduce the inference timesteps from 1000 to 300 or fewer timesteps. These advances enable diffusion models to be efficiently used in clinical practice and allow for more advanced architectures such as Diffusion Transformers (DiT-3D) (Peebles and Xie, 2023, Mo et al., 2023) by reducing computational requirements.

In this paper, we propose the 3D latent rectified flow (T1C-RFlow) model for the prediction of tumor-specific T1-constrast enhancement from multiparametric (T1 and T2-FLAIR) volumes. We perform the training in a compressed latent space representation and implement the rectified flow noise scheduler, improving performance while reducing the computation requirements for training and inference.

The impact and contributions of this paper can be summarized as follows:

- Model training and inference are performed in the latent space of Monai Medical AI for Synthetic Imaging's (MAISI's) pretrained 3D variational autoencoder (VAE) (Guo et al., 2024), reducing the spatial dimensions from 1×256×256×192 to 4×64×64×48 and allowing the entire volume to be captured by the network.
- An RFlow noise scheduler is used which outperforms DDPM in fewer timesteps (200 vs 1000 timesteps). Our model's denoising time (6.9 s/volume, 200 steps) is significantly reduced compared to conventional 3D DDPM models in both latent space (37.7s, 1000 steps) and patch-based in image space (4.3 hr/volume).
- We train our model on T1w and T2-FLAIR images from 3 Brain Tumor Segmentation (BraTS) 2024 challenge datasets (de Verdier et al., 2024) - glioma (GLI; 1480 patients), meningioma (MEN; 1141 patients), and metastases (MET; 1475 patients), so that our model is generalizable to several disease types.

## 2. METHODS

### 2.1 Data Acquisition and Preprocessing

A multi-institutional cohort was assembled from the BraTS 2024 challenge (de Verdier et al., 2024) comprised of glioma (GLI; 1,480), meningioma (MEN; 1,141), and metastasis (MET; 1,475) cases with co-registered T1w, T2-FLAIR, and T1C volumes. The provided MRI volumes were converted from DICOM to NIfTI format and skull stripped using HD-BET. (Isensee et al., 2019) The brain-extracted T1, T1C, T2, and T2-FLAIR sequences were coregistered to the Linear Symmetrical MNI Atlas with affine registration. Additionally, we resampled all scans to 1 mm isotropic resolution (256×256×192 voxels). Intensities were normalized to [-1,1] across the input volume. We randomly split subjects into train, test, and validation splits of 2860, 614, and 612 patients respectively while maintaining the relative proportions of GLI, MEN, and MET cases in each split.

### 2.2 VAE

A VAE learns an encoder $q_\varphi(z|x)$ and decoder $p_\theta(\hat{x}|z)$ over a latent variable, z, with a simple prior $p(z)=\mathcal{N}(\mathbf{0}, \mathbf{I})$. (Kingma and Welling, 2022) The encoder maps an input x to the parameters of a diagonal-Gaussian posterior, μ(x) and σ(x). A latent sample is then drawn via the reparameterization trick

$$z = \boldsymbol{\mu}(x) + \boldsymbol{\sigma}(x)\boldsymbol{\varepsilon}, \qquad \boldsymbol{\varepsilon} \sim \mathcal{N}(\mathbf{0}, \boldsymbol{I}) \qquad (1)$$

which enables low-variance gradient estimates. The decoder reconstructs $\hat{x}$ from z. VAEs are trained by maximizing the evidence lower bound (ELBO), combining a reconstruction term with a Kullback-Leibler (KL) divergence that regularizes $q_\varphi(z|x)$ toward the Gaussian prior, improving stability for latent-space transport methods.(Asperti and Trentin, 2020)

We employ MAISI's pretrained 3D VAE to obtain compact volumetric latents for each sequence (T1w, T2-FLAIR, T1C). (Guo et al., 2024) Inputs are passed through a UNet-style encoder that yields two 4-channel tensors **μ(x)** and **σ(x)** ∈ $\mathbb{R}^{4\times64\times64\times48}$ from a 256×256×192 volume (i.e., 1/4 spatial resolution per axis; 64× fewer voxels). Stochastic latents are formed via (1), and we precompute **μ** and **σ** before training. During training, the T1w and T2-FLAIR conditioning latents are concatenated channelwise, while the T1C target latent serves as the ground truth. During inference, the sT1C latent is decoded by the fixed decoder to reconstruct a full-resolution 3D T1C volume. Operating in this low-dimensional latent space yields substantial memory and time savings while preserving the anatomical details of the tumor region.

## 2.3 Rectified Flow Diffusion U-Net

RFlow casts generation as learning a time-conditioned velocity field $v_\theta(z, t, c)$ that transports a simple noise distribution to the data distribution along a deterministic ordinary differential equation (Liu et al., 2022b) Unlike score-based DDPMs, which require stochastic reverse-time integration with many denoising steps, RFlow learns the drift directly, enabling deterministic sampling with substantially fewer steps.(Lee et al., 2024) Let $z_0$ denote a clean latent (T1C target) and $z_1 \sim \mathcal{N}(\mathbf{0}, \mathbf{I})$ denote a noise latent. With a monotone schedule $\alpha(t)$ such that $\alpha(0) = 0$ and $\alpha(1) = 1$, we define the linear path (2) and the corresponding ground-truth velocity (3):

$$z_t = \big(1 - \alpha(t)\big) z_0 + \alpha(t) z_1 \qquad (2)$$

$$v^*(z_t, t) = \frac{d}{dt} z_t = \alpha'(t)(z_1 - z_0) \qquad (3)$$

We train a 3D diffusion UNet with a symmetrical encoder and decoder containing [128, 128, 256] channels at each layer and 2 residual blocks per layer to approximate $v_\theta(z_t, t, c)$ with an $l_1$ transport loss against the analytical target above, using timesteps drawn from the RFlow scheduler (1000 training timesteps) and conditioning vector, $c$, formed by channel-wise concatenation of the T1W and T2-FLAIR latents (Figure 1). The network input is the concatenation, $[z_t || z_{T1W} || z_{FLAIR}]$ (12×64×64×48) where || denotes the channel-wise concatenation operator. Sampling integrates the learned ODE deterministically from t=1 to t=0 using a fixed-step solver with K steps (default K=200):

$$\mathcal{L}_{RFLOW}(\theta) = \mathbb{E}_{z_0, z_1, t}[||v_\theta(z_t, t, c) - v^*(z_t, t)||_1] \qquad (4)$$

$$\frac{dz}{dt} = v_\theta(z, t, c), \qquad z(1) \sim \mathcal{N}(\mathbf{0}, \mathbf{I}), \qquad t: 1 \to 0 \tag{5}$$

$$z^{k-1} = z^k - \Delta t\, v_\theta(z^k, t_k, c), \quad k = K, \dots, 1, \quad \Delta t = \frac{1}{k} \tag{6}$$

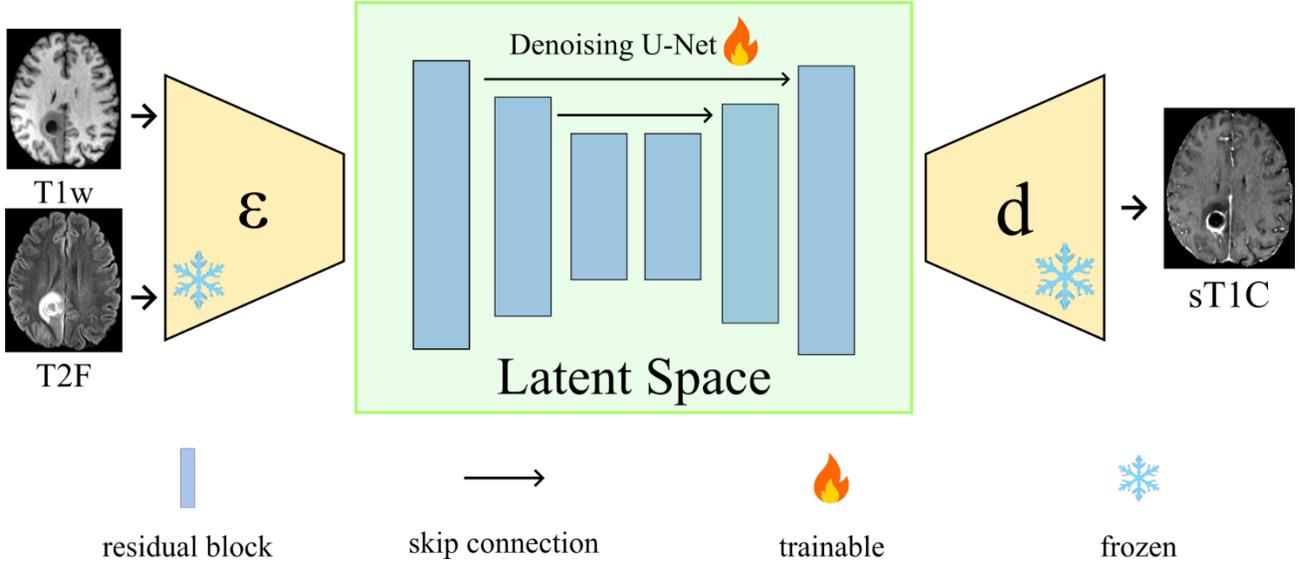

**Figure 1.** Schematic flow diagram of T1C-RFlow. MAISI's pretrained autoencoder compresses and decompresses the input T1w and T2-FLAIR images latent space representations. These latents together with a timestep embedding and noised vector are input into a rectified flow denoising U-Net which is trained to output the corresponding T1C latent.

**2.5 Implementation Details**

The T1C-RFlow model and competing methods were trained with the AdamW optimizer (Loshchilov and Hutter, 2019) (learning rate $1 \times 10^{-5}$, $\beta_1 = 0.9$, $\beta_2 = 0.999$, weight-decay $1 \times 10^{-4}$) for 100 epochs with a batch size of 4 on a single NVIDIA A6000 ADA (48 GB). The rectified flow scheduler implemented 1000 training timesteps and a logit-normal sampling distribution(Esser et al., 2024) based on the latent size ($64 \times 64 \times 48$) with 3 spatial dimensions. Competing methods (DDPM and Pix2pix) were trained according to the hyperparameters and learning schedule as proposed in the original works.(Isola et al., 2018, Ho et al., 2020) The DiT-3D model was modified to work on medical images as opposed to point clouds in the original implementation (Mo et al., 2023) and to use a rectified flow noise scheduler. During inference, T1C-RFlow and DiT-3D both used 200 timesteps while DDPM used 1000 timesteps. The generated latents from all methods were then translated to image space with the VAE's decoder for evaluation.

**2.6 Validation and Evaluation**

We evaluated the final performance on the test set (N = 614) of the holdout test. All methods were evaluated on the full 3D decoded image-space volumes normalized from [-1,1]. For each patient, quantitative accuracy was computed over the whole MRI volume and summarized as mean ± std across the test cohort and further stratified by disease type (GLI; N = 239, MEN; N = 162, MET; N = 213). We report normalized mean-squared error (NMSE) (Safari et al., 2024), peak signal-to-noise ratio (PSNR) (Safari et al., 2025), normalized cross-correlation (NCC) (Qiu et al., 2021), and structural similarity (SSIM) (Eidex et al., 2024). The SSIM was formulated with the stabilizing constants, $C_1$ and $C_2$, set to $C_1 = (0.01L)^2$ and $C_2 = (0.03L)^2$ where L is the dynamic range (L = 2) In addition to calculating metrics for the whole brain volume, we also calculated metrics for the tumor region. We defined the tumor region by creating a bounding box around the provided segmentation maps to define a bounding box with a padding of 5 voxels and calculated the metrics within this bounding box. Since some segmentation maps were not provided, a subset of 545 patients with segmentation maps was used to evaluate the tumor region. Since Of Ofasdasd asdasdasdasdasdasdasd rtrrrewdqasdasdasdasdasdsasddasd TaaaaaaaaaaaaaaaaNMSE quantifies the normalized residual energy and increases monotonically with error (lower is better); PSNR restates the mean-squared error on a logarithmic decibel scale relative to the dynamic range and tends to align with perceived fidelity (higher is better). NCC measures linear agreement in structure while being insensitive to global intensity shifts and global scaling (higher is better); and SSIM evaluates luminance, contrast, and structural similarity to better align with human perception (higher is better).

For statistical analysis, per-patient metric values were compared using a two-sided Welch's t-test (unequal variances) within each cohort (overall and across each dataset) (Delacre et al., 2017). A significance level of α=0.05 and $p<0.05$ was considered statistically significant.

## 3. RESULTS

The proposed T1C-RFlow model demonstrated superior performance in synthesizing T1C MRI compared to existing state-of-the-art methods. The T1C-RFlow model achieved the highest fidelity metrics for whole brain volumes on the test set, with noticeably improved mean NMSE, PSNR, and NCC (Table 1). In particular, the proposed method improved average NMSE by 25% compared to the next best method (DDPM). All improvements were statistically significant except for SSIM when compared to pix2pix. These results remained consistent when stratified by dataset (GLI, MEN, and MET) although all methods produced quantitatively worse sT1C for metastases compared to glioma and meningioma. This may be due to lower quality T1C images in this dataset. T1C For the tumor region, T1C-RFlow again had the highest performance on the evaluation metrics with most differences being statistically significant ($p<.05$). The GLI dataset contained the most heterogeneous, complex tumors which may account for why the methods had the lowest quantitative results (Table 2(.

We also compared model inference time on an A6000 ADA GPU (Table 2) and found the total time of the T1C-RFlow model to be a reasonable 9.7 seconds (1.2s encode + 6.9s denoising + 1.6s decoding) especially compared to conventional 3D patch-based DDPM methods (4.3 hours). Pix2pix had the fastest inference time since the latent prediction happens in a single step compared to 200 steps for T1C-RFlow and 1000 steps for the tested DDPM model. However, sT1C generated by pix2pix was suboptimal in terms of the image similarity/quality metrics. Compressing the images with a VAE resulted in minimal visual differences and was verified quantitively with the T1W, T2-FLAIR, and T1-contrast volumes having PSNR values near 35 dB (Table 3). Finally, we found that a multimodal approach (T1w + T2-FLAIR) was significantly better ($p<.01$) than T1w or T2-FLAIR alone by running model inference with and without zeroing the input tensors from each sequence (Table 4). Visually, the sT1C images generated by T1C-RFlow are closest to the real contrast-enhanced scans, although regions with contrast show slight visual differences and the overall brightness of T1C and sT1C of the proposed method can be in the are very different especially with low quality ground truth T1C scans (Figure 2; rows (e)-(g)). T1C-RFlow outperforms competing methods for glioma (Figure 2; Rows (a)-(d)) meningioma (Figure 2; rows (e) and (f)) and metastases (Figure 2; rows (g) and (h)).

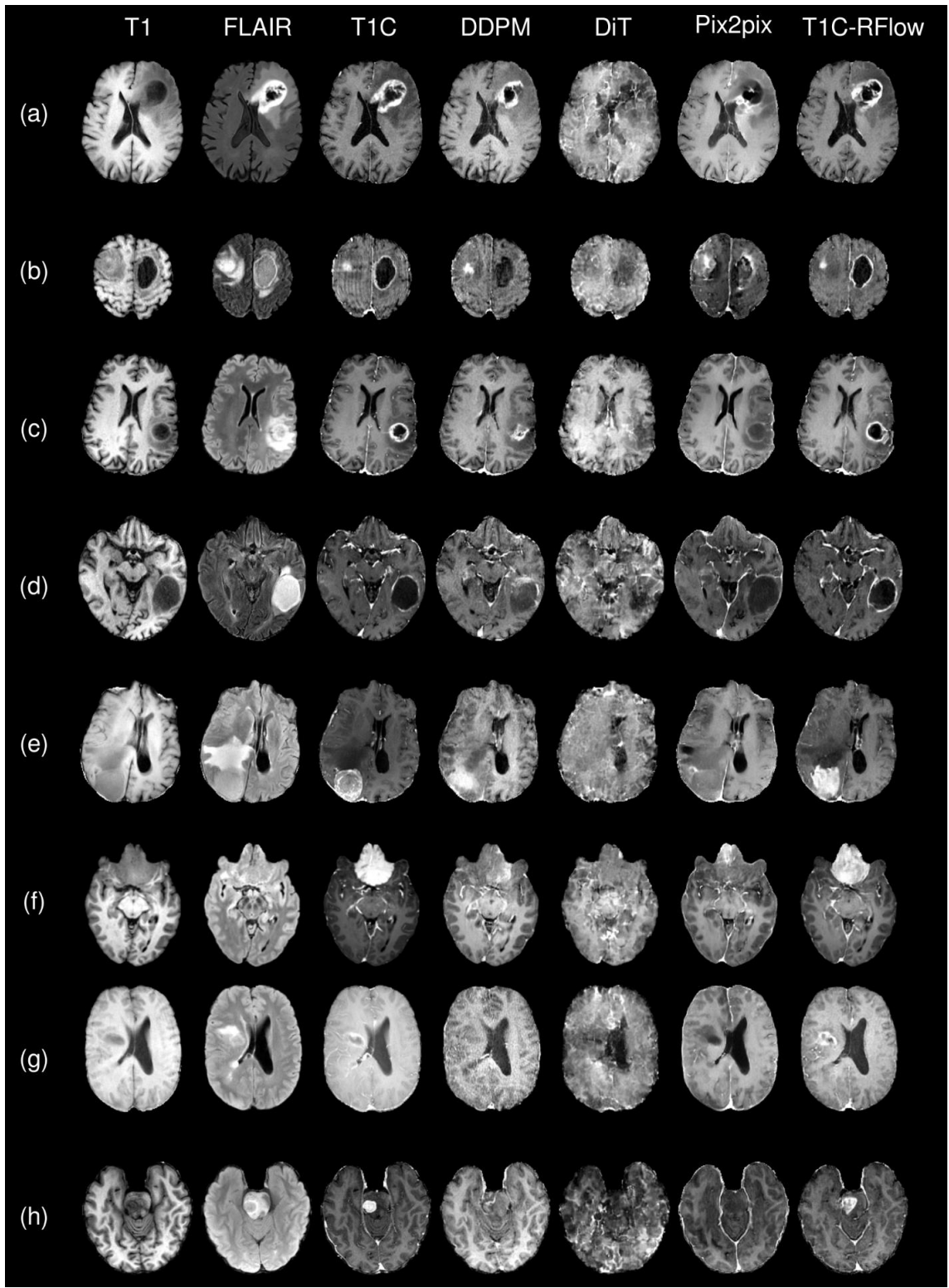

**Figure 2.** Example sT1C compared to ground truth slices generated from DDPM, DiT, Pix2pix, and T1C-RFlow together with the input T1w and T2-FLAIR MRIs. Rows (a)-(d) are glioma (BraTS-GLI); rows (e) and (f) are meningioma (BraTS-MEN); rows (g) and (h) are metastases (BraTS-MET).

**Table 1.** Quantitative performance for the whole brain volume of the proposed method compared to state-of-the-art methods for the GLI, MEN, and MET datasets as well as averaged together. All methods are 3D and trained in the same latent space as the proposed method. P-value metrics are compared against T1C-RFlow's performance.

| | | NMSE[↓] | | PSNR (dB) [↑] | | NCC[↑] | | SSIM[↑] | |
|---|---|---|---|---|---|---|---|---|---|
| | | Mean ± Std. | p-value | Mean ± Std. | p-value | Mean ± Std. | p-value | Mean ± Std. | p-value |
| All | Proposed (T1C-RFlow) | **0.063 ± 0.067** | - | **29.7 ± 3.0** | - | **0.957 ± 0.018** | - | **0.925 ± 0.054** | - |
| | DDPM | 0.084 ± 0.078 | <.001 | 28.3 ± 3.0 | <.001 | 0.951 ± 0.021 | <.001 | 0.903 ± 0.059 | <.001 |
| | Pix2pix | 0.091 ± 0.250 | 0.007 | 28.4 ± 2.9 | <.001 | 0.933 ± 0.030 | <.001 | **0.923 ± 0.051** | 0.546 |
| | DiT-3D | 0.118 ± 0.074 | <.001 | 26.5 ± 3.0 | <.001 | 0.922 ± 0.030 | <.001 | 0.874 ± 0.077 | <.001 |
| GLI | Proposed (T1C-RFlow) | **0.044 ± 0.047** | - | **29.9 ± 2.9** | - | **0.960 ± 0.015** | x | **0.935 ± 0.025** | - |
| | DDPM | 0.064 ± 0.045 | <.001 | 28.2 ± 2.9 | <.001 | 0.957 ± 0.016 | 0.036 | 0.910 ± 0.041 | <.001 |
| | Pix2pix | 0.057 ± 0.041 | 0.002 | 28.7 ± 2.7 | <.001 | 0.942 ± 0.029 | <.001 | 0.931 ± 0.027 | 0.109 |
| | DiT-3D | 0.096 ± 0.055 | <.001 | 26.4 ± 2.9 | <.001 | 0.928 ± 0.028 | <.001 | 0.885 ± 0.047 | <.001 |
| MEN | Proposed (T1C-RFlow) | **0.046 ± 0.029** | - | **29.9 ± 2.7** | - | **0.961 ± 0.014** | - | **0.937 ± 0.021** | - |
| | DDPM | 0.064 ± 0.037 | <.001 | 28.3 ± 2.7 | <.001 | 0.956 ± 0.018 | 0.005 | 0.913 ± 0.035 | <.001 |
| | Pix2pix | 0.058 ± 0.034 | <.001 | 28.7 ± 2.5 | <.001 | 0.942 ± 0.018 | <.001 | 0.936 ± 0.022 | 0.45 |
| | DiT-3D | 0.104 ± 0.062 | <.001 | 26.2 ± 2.8 | <.001 | 0.930 ± 0.023 | <.001 | 0.881 ± 0.053 | <.001 |
| MET | Proposed (T1C-RFlow) | **0.098 ± 0.088** | - | **29.4 ± 3.3** | - | **0.949 ± 0.020** | - | 0.905 ± 0.082 | - |
| | DDPM | 0.122 ± 0.111 | 0.013 | 28.3 ± 3.2 | <.001 | 0.941 ± 0.024 | <.001 | 0.886 ± 0.084 | 0.02 |
| | Pix2pix | 0.155 ± 0.412 | 0.05 | 27.9 ± 3.4 | <.001 | 0.918 ± 0.033 | <.001 | **0.905 ± 0.075** | 0.94 |
| | DiT-3D | 0.154 ± 0.085 | <.001 | 26.8 ± 3.3 | <.001 | 0.911 ± 0.032 | <.001 | 0.855 ± 0.110 | <.001 |

**Table 2.** Quantitative performance for the tumor region of the proposed method compared to state-of-the-art methods for the GLI, MEN, and MET datasets as well as averaged together.

| | | NMSE[↓] | | PSNR (dB) [↑] | | NCC[↑] | | SSIM[↑] | |
|---|---|---|---|---|---|---|---|---|---|
| | | Mean ± Std. | - | Mean ± Std. | p-value | Mean ± Std. | p-value | Mean ± Std. | p-value |
| All | Proposed (T1C-RFlow) | **0.074 ± 0.054** | - | **21.7 ± 5.8** | - | **0.486 ± 0.290** | - | **0.570 ± 0.170** | - |
| | DDPM | 0.126 ± 0.566 | 0.035 | 20.4 ± 5.0 | <.001 | 0.435 ± 0.302 | 0.004 | 0.504 ± 0.180 | <.001 |
| | Pix2pix | 0.135 ± 0.610 | 0.021 | 20.0 ± 5.5 | <.001 | 0.403 ± 0.310 | <.001 | 0.549 ± 0.171 | 0.039 |
| | DiT-3D | 0.151 ± 0.169 | <.001 | 18.5 ± 4.8 | <.001 | 0.309 ± 0.301 | <.001 | 0.438 ± 0.166 | <.001 |
| GLI | Proposed (T1C-RFlow) | **0.090 ± 0.047** | - | **19.5 ± 2.9** | - | **0.282 ± 0.196** | - | **0.462 ± 0.127** | - |
| | DDPM | 0.126 ± 0.116 | <.001 | 18.6 ± 2.9 | 0.001 | 0.219 ± 0.210 | 0.001 | 0.407 ± 0.124 | <.001 |
| | Pix2pix | 0.113 ± 0.070 | <.001 | 18.7 ± 2.9 | 0.007 | 0.194 ± 0.232 | <.001 | 0.476 ± 0.137 | 0.274 |
| | DiT-3D | 0.162 ± 0.092 | <.001 | 17.1 ± 3.4 | <.001 | 0.122 ± 0.185 | <.001 | 0.389 ± 0.116 | <.001 |
| MEN | Proposed (T1C-RFlow) | **0.061 ± 0.058** | - | **24.8 ± 8.7** | - | **0.658 ± 0.277** | - | **0.699 ± 0.182** | - |
| | DDPM | 0.083 ± 0.067 | 0.004 | 23.0 ± 7.2 | 0.057 | 0.604 ± 0.291 | 0.106 | 0.641 ± 0.202 | 0.011 |
| | Pix2pix | 0.091 ± 0.121 | 0.007 | 23.2 ± 8.2 | 0.115 | 0.623 ± 0.271 | 0.271 | 0.672 ± 0.191 | 0.223 |
| | DiT-3D | 0.130 ± 0.100 | <.001 | 20.7 ± 6.7 | <.001 | 0.491 ± 0.307 | <.001 | 0.564 ± 0.208 | <.001 |
| MET | Proposed (T1C-RFlow) | **0.067 ± 0.054** | - | **21.8 ± 4.1** | - | **0.589 ± 0.248** | - | **0.596 ± 0.119** | - |
| | DDPM | 0.159 ± 0.957 | 0.193 | 20.5 ± 3.6 | <.001 | 0.554 ± 0.247 | 0.172 | 0.509 ± 0.142 | <.001 |

| | | | | | | | |
|---|---|---|---|---|---|---|---|
| Pix2pix | 0.195 ± 1.032 | 0.093 | 18.9 ± 4.1 | <.001 | 0.474 ± 0.263 | <.001 | 0.538 ± 0.132 | <.001 |
| DiT-3D | 0.154 ± 0.257 | <.001 | 18.4 ± 3.7 | <.001 | 0.383 ± 0.285 | <.001 | 0.398 ± 0.126 | <.001 |

**Table 2.** Inference times on an A6000 ADA GPU. The proposed method takes less than 10 seconds per volume compared to over 4 hours using a conventional patch based DDPM diffusion model.

| | Timesteps | Inference Time |
|---|---|---|
| Autoencoder (Encode) | x | 1.2 s |
| Autoencoder (Decode) | x | 1.6 s |
| Pix2pix | x | 74.0 ms |
| DDPM (Latent) | 1000 | 37.7 s |
| DDPM (Patches) | 1000 | 4.3 hr |
| DiT-3D | 200 | 11.9 s |
| Proposed (T1C-RFlow) | 200 | 6.9 s |

**Table 3.** Quantitative performance of MAISI's VAE ability to accurately encode and decode T1w, T2-FLAIR, and T1C sequences compared to the original input image.

| Modality | NMSE(×$10^{-2}$) [↓] | PSNR (dB) [↑] | SSIM[↑] |
|---|---|---|---|
| **T1w** | 0.63 ± 0.25 | 35.3 ± 2.0 | 0.979 ± 0.006 |
| **T2-FLAIR** | 1.11 ± 0.58 | 34.8 ± 1.9 | 0.972 ± 0.008 |
| **T1C** | 1.63 ± 1.11 | 35.2 ± 1.5 | 0.973 ± 0.008 |

**Table 4.** Ablation studies of the proposed method with T1w, T2-FLAIR, and both T1w and T2-FLAIR MRI. Our model achieved the best performance using information from both sequences together.

| | NMSE[↓] | | PSNR (dB) [↑] | | NCC[↑] | | SSIM[↑] | |
|---|---|---|---|---|---|---|---|---|
| | Mean + Std. | p-value | Mean + Std. | p-value | Mean + Std. | p-value | Mean + Std. | p-value |
| T1w + T2-FLAIR | **0.063 ± 0.067** | x | **29.7 ± 3.0** | x | **0.957 ± 0.017** | x | **0.925 ± 0.054** | x |
| T1w only | 0.074 ± 0.071 | 0.008 | 29.0 ± 3.0 | <.001 | 0.946 ± 0.024 | <.001 | 0.916 ± 0.054 | 0.004 |
| T2-FLAIR only | 0.117 ± 0.079 | <.001 | 26.4 ± 2.5 | <.001 | 0.892 ± 0.034 | <.001 | 0.881 ± 0.060 | <.001 |

## 4. DISCUSSION

In this study, we demonstrate the T1C-RFlow model quantitatively outperforms the state-of-the art DDPM, the GAN-based Pix2pix, and diffusion-transformer-based DiT-3D models while generating sT1C volumes in less than 10 seconds. By training our model using T1w and T2-FLAIR images from 3 Brain Tumor Segmentation (BraTS) 2024 challenge datasets (de Verdier et al., 2024) (GLI, MEN, and MET) we make our model generalizable to several disease types. Further development of our approach may permit a practical method for capturing the entire 3D image context for contrast-agent-free MRI for brain tumors, eliminating GBCA-associated risk and simplifying clinical protocols.

We use a 3D LDM for synthesizing T1C from multiparametric MRI. Conventional 2D slice-by-slice models struggle to capture volumetric context leading to inconsistent enhancement across slices.

Patch-based 3D models must similarly limit their context to a portion of the MRI volume and incorporate computationally expensive overlapping tile strategies to avoid visual artifacts due to discrepancies between tiles. Our 3D latent diffusion framework (T1C-RFlow) addresses these issues by operating in a compressed latent space, which drastically reduces memory requirements and captures the entire 3D volume at once.

Our design builds upon insights from recent works. Eidex *et al.* improved synthetic T1C quality using a vision Transformer (ViT) conditioned on tumor segmentation maps. (Eidex et al., 2025a) Ma *et al.* leveraged a 2D latent diffusion model with ControlNet guidance to synthesize T1C from T2-FLAIR, achieving superior image quality and aiding tumor delineation. (Ma et al., 2025) In addition, Li *et al* showed the advantages of a multiparametric transformer-based approach and integrating context from the tumor region. (Li et al., 2025b) However, all studies were designed for 2D axial slices and may struggle with long-range 3D context. In addition, these studies did not investigate generalizability to different disease types.

Qualitatively, our method captures fine anatomical details and the contrast-enhancing tumor region much more faithfully than baseline methods (Table 2) and may even surpass or restore detail to low-quality ground truth T1C volumes (e.g. patient motion, artifacts, variability in contrast agent administration) (Figure 2). We identify several factors that may have affected this result. Compared to GAN-based models like pix2pix, diffusion models have demonstrated improved training stability and image quality at the cost of increased inference time. In addition, compared to the original DDPM formulation, rectified flow methods often converge more quickly and produce higher quality results than DDPM in fewer timesteps by traversing a more direct path from noised to target latent space. Although powerful, 3D ViT based methods must compress the 4D source latent into a 1D tensor of image patches before performing the self-attention operation. (Dosovitskiy et al., 2020, Eidex et al., 2025b) The spatial relationships are retained through learnable position embeddings, but this task can be challenging and lead to unstable training.

We note several limitations of this study. First, our study was limited to brain tumor cohorts from the BraTS 2024 challenge, so performance on other neurological diseases or imaging protocols remains to be validated. Second, while our model implicitly captured tumor enhancement patterns, it did not use explicit tumor masks or structural priors during training. While technically possible, this was avoided since MAISI's pretrained VAE was not trained directly on segmentation maps so would require extensive finetuning to accurately reconstruct the segmentation maps. This may help in difficult cases with large, heterogeneous tumor volumes. Scaling T1C-RFlow toward a foundation model trained on larger, multi-institutional datasets could further improve robustness and generalizability. (Wang et al., 2025b, Wang et al., 2025a).

Future work will investigate developing a foundation model for generating multiple MR sequences compared to only T1-contrast on an even larger multimodal dataset. We are also interested in achieving higher performance with ViT, Swin transformer, (Liu et al., 2021, Pan et al., 2023) and highly optimized convolution based models which can situationally outperform ViT models. (Liu et al., 2022c) Finally, we are confident that integrating segmentation map information, either directly or by prediction, into the sT1C generation pipeline will improve tumor region performance.

## 5. CONCLUSION

This study introduces T1C-RFlow, an efficient 3D latent diffusion model for synthesizing high-quality T1-contrast-like scans from multiparametric MRI for glioma, meningioma, and metastasis patients while making predictions in much less time than previous diffusion models. Our proposed method shows promising potential as a contrast-agent-free MRI alternative for imaging characterization of brain tumors, eliminating GBCA-associated risk and simplifying clinical protocols.

**ACKNOWLEDGMENTS**

This research is supported in part by the National Cancer Institute of the National Institutes of Health under Award Numbers R01CA272991, R01DE33512 and R01CA272755.

**Disclosures**

The author declares no conflicts of interest.

**Reference:**

ASPERTI, A. & TRENTIN, M. 2020. Balancing Reconstruction Error and Kullback-Leibler Divergence in Variational Autoencoders. *IEEE Access,* 8**,** 199440-199448.
BECKER, E., PANDIT, P., RANGAN, S. & FLETCHER, A. K. 2022. Instability and Local Minima in GAN Training with Kernel Discriminators. *arXiv [cs.LG]*.
BIEDER, F., WOLLEB, J., DURRER, A., SANDKÜHLER, R. & CATTIN, P. C. 2024. Memory-Efficient 3D Denoising Diffusion Models for Medical Image Processing. *arXiv [cs.CV]*.
CHANG, C.-W., PENG, J., SAFARI, M., SALARI, E., PAN, S., ROPER, J., QIU, R. L. J., GAO, Y., SHU, H.-K. & MAO, H. 2024. High-resolution MRI synthesis using a data-driven framework with denoising diffusion probabilistic modeling. *Physics in Medicine & Biology,* 69**,** 045001.
DE VERDIER, M. C., SALUJA, R., GAGNON, L., LABELLA, D., BAID, U., TAHON, N. H., FOLTYN-DUMITRU, M., ZHANG, J., ALAFIF, M., BAIG, S., CHANG, K., D'ANNA, G., DEPTULA, L., GUPTA, D., HAIDER, M. A., HUSSAIN, A., IV, M., KONTZIALIS, M., MANNING, P., MOODI, F., NUNES, T., SIMON, A., SOLLMANN, N., VU, D., ADEWOLE, M., ALBRECHT, J., ANAZODO, U., CHAI, R., CHUNG, V., FAGHANI, S., FARAHANI, K., KAZEROONI, A. F., IGLESIAS, E., KOFLER, F., LI, H., LINGURARU, M. G., MENZE, B., MOAWAD, A. W., VELICHKO, Y., WIESTLER, B., ALTES, T., BASAVASAGAR, P., BENDSZUS, M., BRUGNARA, G., CHO, J., DHEMESH, Y., FIELDS, B. K. K., GARRETT, F., GASS, J., HADJIISKI, L., HATTANGADI-GLUTH, J., HESS, C., HOUK, J. L., ISUFI, E., LAYFIELD, L. J., MASTORAKOS, G., MONGAN, J., NEDELEC, P., NGUYEN, U., OLIVA, S., PEASE, M. W., RASTOGI, A., SINCLAIR, J., SMITH, R. X., SUGRUE, L. P., THACKER, J., VIDIC, I., VILLANUEVA-MEYER, J., WHITE, N. S., ABOIAN, M., CONTE, G. M., DALE, A., SABUNCU, M. R., SEIBERT, T. M., WEINBERG, B., ABAYAZEED, A., HUANG, R., TURK, S., RAUSCHECKER, A. M., FARID, N., VOLLMUTH, P., NADA, A., BAKAS, S., CALABRESE, E. & RUDIE, J. D. 2024. The 2024 Brain Tumor Segmentation (BraTS) Challenge: Glioma Segmentation on Post-treatment MRI. *arXiv [cs.CV]*.

DELACRE, M., LAKENS, D. & LEYS, C. 2017. Why psychologists should by default use Welch's t-test instead of Student's t-test. *International Review of Social Psychology,* 30**,** 92-101.
DOSOVITSKIY, A., BEYER, L., KOLESNIKOV, A., WEISSENBORN, D., ZHAI, X., UNTERTHINER, T., DEHGHANI, M., MINDERER, M., HEIGOLD, G. & GELLY, S. 2020. An image is worth 16x16 words: Transformers for image recognition at scale. *arXiv preprint arXiv:2010.11929*.
EIDEX, Z., DING, Y., WANG, J., ABOUEI, E., QIU, R. L. J., LIU, T., WANG, T. & YANG, X. 2023. Deep Learning in MRI-guided Radiation Therapy: A Systematic Review. *ArXiv*.
EIDEX, Z., SAFARI, M., QIU, R. L. J., YU, D. S., SHU, H. K., MAO, H. & YANG, X. 2025a. T1-contrast enhanced MRI generation from multi-parametric MRI for glioma patients with latent tumor conditioning. *Med Phys,* 52**,** 2064-2073.
EIDEX, Z., SAFARI, M., WYNNE, J., QIU, R. L. J., WANG, T., VIAR-HERNANDEZ, D., SHU, H. K., MAO, H. & YANG, X. 2025b. Deep learning based apparent diffusion coefficient map generation from multi-parametric MR images for patients with diffuse gliomas. *Medical Physics,* 52**,** 847-855.
EIDEX, Z., WANG, J., SAFARI, M., ELDER, E., WYNNE, J., WANG, T., SHU, H. K., MAO, H. & YANG, X. 2024. High-resolution 3T to 7T ADC map synthesis with a hybrid CNN-transformer model. *Med Phys,* 51**,** 4380-4388.
ESSER, P., KULAL, S., BLATTMANN, A., ENTEZARI, R., MÜLLER, J., SAINI, H., LEVI, Y., LORENZ, D., SAUER, A., BOESEL, F., PODELL, D., DOCKHORN, T., ENGLISH, Z., LACEY, K., GOODWIN, A., MAREK, Y. & ROMBACH, R. 2024. Scaling Rectified Flow Transformers for High-Resolution Image Synthesis. *arXiv [cs.CV]*.
GUO, P., ZHAO, C., YANG, D., XU, Z., NATH, V., TANG, Y., SIMON, B., BELUE, M., HARMON, S., TURKBEY, B. & XU, D. 2024. MAISI: Medical AI for Synthetic Imaging. *arXiv [eess.IV]*.
HO, J., JAIN, A. & ABBEEL, P. 2020. Denoising Diffusion Probabilistic Models. *arXiv [cs.LG]*.
IBRAHIM, M. A., HAZHIRKARZAR, B. & DUBLIN, A. B. 2018. Gadolinium magnetic resonance imaging.
ISENSEE, F., SCHELL, M., PFLUEGER, I., BRUGNARA, G., BONEKAMP, D., NEUBERGER, U., WICK, A., SCHLEMMER, H. P., HEILAND, S., WICK, W., BENDSZUS, M., MAIER-HEIN, K. H. & KICKINGEREDER, P. 2019. Automated brain extraction of multisequence MRI using artificial neural networks. *Hum Brain Mapp,* 40**,** 4952-4964.
ISOLA, P., ZHU, J.-Y., ZHOU, T. & EFROS, A. A. 2018. Image-to-Image Translation with Conditional Adversarial Networks. *arXiv [cs.CV]*.
IYAD, N., M, S. A., ALKHATIB, S. G. & HJOUJ, M. 2023. Gadolinium contrast agents- challenges and opportunities of a multidisciplinary approach: Literature review. *Eur J Radiol Open,* 11**,** 100503.
KINGMA, D. P. & WELLING, M. 2022. Auto-Encoding Variational Bayes. *arXiv [stat.ML]*.
LEE, S., LIN, Z. & FANTI, G. 2024. Improving the Training of Rectified Flows. *arXiv [cs.CV]*.
LI, L., SU, C., GUO, Y., ZHANG, H., LIANG, D. & SHANG, K. 2025a. Interactive Gadolinium-Free MRI Synthesis: A Transformer with Localization Prompt Learning. *arXiv [eess.IV]*.
LI, L., SU, C., GUO, Y., ZHANG, H., LIANG, D. & SHANG, K. 2025b. Interactive Gadolinium-Free MRI Synthesis: A Transformer with Localization Prompt Learning. *arXiv preprint arXiv:2503.01265*.
LIU, L., REN, Y., LIN, Z. & ZHAO, Z. 2022a. Pseudo Numerical Methods for Diffusion Models on Manifolds. *arXiv [cs.CV]*.
LIU, X., GONG, C. & LIU, Q. 2022b. Flow Straight and Fast: Learning to Generate and Transfer Data with Rectified Flow. *arXiv [cs.LG]*.
LIU, Z., LIN, Y., CAO, Y., HU, H., WEI, Y., ZHANG, Z., LIN, S. & GUO, B. 2021. Swin Transformer: Hierarchical Vision Transformer using Shifted Windows. *arXiv [cs.CV]*.
LIU, Z., MAO, H., WU, C.-Y., FEICHTENHOFER, C., DARRELL, T. & XIE, S. 2022c. A ConvNet for the 2020s. *arXiv [cs.CV]*.
LOSHCHILOV, I. & HUTTER, F. 2019. Decoupled Weight Decay Regularization. *arXiv [cs.LG]*.

MA, X., MA, Y., WANG, Y., LI, C., LIU, Y., CHEN, X., DAI, J., BI, N. & MEN, K. 2025. Contrast-enhanced image synthesis using latent diffusion model for precise online tumor delineation in MRI-guided adaptive radiotherapy for brain metastases. *Physics in Medicine and Biology*.

MO, S., XIE, E., CHU, R., YAO, L., HONG, L., NIEßNER, M. & LI, Z. 2023. DiT-3D: Exploring Plain Diffusion Transformers for 3D Shape Generation. *arXiv [cs.CV]*.

PAN, S., ABOUEI, E., WYNNE, J., CHANG, C.-W., WANG, T., QIU, R. L. J., LI, Y., PENG, J., ROPER, J., PATEL, P., YU, D. S., MAO, H. & YANG, X. 2024. Synthetic CT generation from MRI using 3D transformer-based denoising diffusion model. *Medical Physics,* 51**,** 2538-2548.

PAN, S., WANG, T., QIU, R. L. J., AXENTE, M., CHANG, C. W., PENG, J., PATEL, A. B., SHELTON, J., PATEL, S. A., ROPER, J. & YANG, X. 2023. 2D medical image synthesis using transformer-based denoising diffusion probabilistic model. *Phys Med Biol,* 68.

PEEBLES, W. & XIE, S. 2023. Scalable Diffusion Models with Transformers. *arXiv [cs.CV]*.

QIU, R. L. J., LEI, Y., SHELTON, J., HIGGINS, K., BRADLEY, J. D., CURRAN, W. J., LIU, T., KESARWALA, A. H. & YANG, X. 2021. Deep learning-based thoracic CBCT correction with histogram matching. *Biomed Phys Eng Express,* 7.

ROMBACH, R., BLATTMANN, A., LORENZ, D., ESSER, P. & OMMER, B. 2022. High-Resolution Image Synthesis with Latent Diffusion Models. *arXiv [cs.CV]*.

ROOZPEYKAR, S., AZIZIAN, M., ZAMANI, Z., FARZAN, M. R., VESHNAVEI, H. A., TAVOOSI, N., TOGHYANI, A., SADEGHIAN, A. & AFZALI, M. 2022. Contrast-enhanced weighted-T1 and FLAIR sequences in MRI of meningeal lesions. *Am J Nucl Med Mol Imaging,* 12**,** 63-70.

SAFARI, M., EIDEX, Z., PAN, S., QIU, R. L. J. & YANG, X. 2025. Self-supervised adversarial diffusion models for fast MRI reconstruction. *Med Phys,* 52**,** 3888-3899.

SAFARI, M., YANG, X., CHANG, C. W., QIU, R. L. J., FATEMI, A. & ARCHAMBAULT, L. 2024. Unsupervised MRI motion artifact disentanglement: introducing MAUDGAN. *Phys Med Biol,* 69.

SHAOYAN, P., ZACH, E., MOJTABA, S., RICHARD, Q. & XIAOFENG, Y. Cycle-guided denoising diffusion probability model for 3D cross-modality MRI synthesis. Proc.SPIE, 2025. 134101W.

SONG, J., MENG, C. & ERMON, S. 2022. Denoising Diffusion Implicit Models. *arXiv [cs.LG]*.

WANG, S., JIN, Z., HU, M., SAFARI, M., ZHAO, F., CHANG, C.-W., QIU, R. L. J., ROPER, J., YU, D. S. & YANG, X. 2025a. Unifying Biomedical Vision-Language Expertise: Towards a Generalist Foundation Model via Multi-CLIP Knowledge Distillation. *arXiv preprint arXiv:2506.22567*.

WANG, S., SAFARI, M., LI, Q., CHANG, C.-W., QIU, R. L. J., ROPER, J., YU, D. S. & YANG, X. 2025b. Triad: Vision Foundation Model for 3D Magnetic Resonance Imaging. *arXiv [cs.CV]*.

ZHOU, Z. & LU, Z. R. 2013. Gadolinium-based contrast agents for magnetic resonance cancer imaging. *Wiley Interdiscip Rev Nanomed Nanobiotechnol,* 5**,** 1-18.